\documentclass[10pt,twocolumn,letterpaper]{article}

\usepackage[pagenumbers]{wacv} 
 \usepackage{bm}
\usepackage{adjustbox}
\usepackage{fontawesome}
\usepackage{pifont} 
\usepackage[table,x11names]{xcolor}
\usepackage{tikz,pgfplots}
\usetikzlibrary{intersections, arrows.meta,positioning,calc, shapes.geometric}
\usepackage{amsthm}
\usepackage{multirow}
\usepackage{colortbl}
\usepackage{float}
\usepackage{algorithm}
\usepackage{algpseudocode}

\newtheorem{assumption}{Assumption}[section]
\newtheorem{proposition}{Proposition}[section]
\newtheorem{corollary}{Corollary}[section]
\newtheorem{remark}{Remark}[section]
\definecolor{wacvblue}{rgb}{0.21,0.49,0.74}
\usepackage[pagebackref,breaklinks,colorlinks,allcolors=wacvblue]{hyperref}

\def\wacvPaperID{*****} 
\def\confName{WACV}
\def\confYear{2026}

\title{Confidence-Calibrating Regularization for Robust Brain MRI Segmentation Under Domain Shift}

\author{Behraj Khan \href{https://orcid.org/0000-0003-0985-9543}{\includegraphics[scale=0.06]{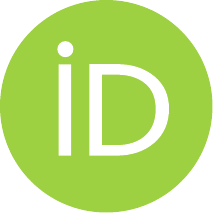}}\\
St. John's University\\
New York, USA\\
{\tt\small behrajkhan@gmail.com}
\and
Tahir Qasim Syed \href{https://orcid.org/0000-0003-0638-9689}{\includegraphics[scale=0.06]{orcid.pdf}}\\
Institute of Business Administration Karachi\\
Pakistan\\
{\tt\small tahirqsyed@gmail.com}
\and
Syed Ahmad Chan Bukhari \href{https://orcid.org/0000-0002-6517-5261}{\includegraphics[scale=0.06]{orcid.pdf}}\\
St. John's University\\
New York, USA\\
{\tt\small bukharis@stjohns.edu}
}

\begin{document}
\maketitle

\abstract{%
The Segment Anything Model (SAM) exhibits strong zero-shot performance on
natural images but suffers from domain shift and overconfidence when applied
to medical volumes. We propose \textbf{CalSAM}, a lightweight adaptation
framework that (i) reduces encoder sensitivity to domain shift via a
\emph{Feature Fisher Information Penalty} (FIP) computed on 3D feature maps
and (ii) penalizes overconfident voxel-wise errors through a
\emph{Confidence Misalignment Penalty} (CMP). The combined loss,
\(
\mathcal{L}_{\mathrm{CalSAM}}
=
\mathcal{L}_{\mathrm{SAM}}
+\lambda_1\mathrm{FIP}
+\lambda_2\mathrm{CMP},
\)
fine-tunes only the mask decoder while keeping SAM's encoders frozen.
On cross-center and scanner-shift evaluations, CalSAM substantially improves
accuracy and calibration: e.g., on the BraTS scanner split
(Siemens$\rightarrow$GE) CalSAM attains
$\mathrm{DSC}=80.1\pm1.3\%$
(vs.\ $74.6\pm2.1\%$ for SAM-FT),
$\mathrm{HD95}=4.6\pm0.7$ mm
(vs.\ $6.3\pm1.0$ mm), and
$\mathrm{ECE}=5.2\pm0.8\%$
(vs.\ $8.6\pm1.3\%$).
On ATLAS-C (motion corruptions) CalSAM reaches
$\mathrm{DSC}=75.9\pm2.0\%$
and $\mathrm{ECE}=5.8\pm0.9\%$.
Ablations show FIP and CMP contribute complementary gains
($p<0.01$), and the Fisher penalty incurs a modest
$\sim15\%$ training-time overhead. CalSAM therefore delivers improved
domain generalization and better-calibrated uncertainty estimates for brain
MRI segmentation, while retaining the computational benefits of freezing
SAM's encoder.
}


\maketitle




\section{Introduction}
\label{sec:intro}
Medical image segmentation is a crucial task for diagnosis, surgical guidance, and treatment planning. Recent advances in foundation models, such as Segment Anything Model (SAM) \cite{kirillov2023segment} demonstrates strong zero-shot generalization capabilities on natural images. However, its effectiveness in medical imaging particularly under domain shifts (e.g. population biases, protocol variations, or scanner differences) remains limited and challenging as it often leads to unreliable segmentation mask \cite{huang2024segment}. SAM and similar pre-trained models often produce highly confident but incorrect predictions in such scenarios, raising concerns about trustworthiness and deployment longevity of medical pre-trained models like SAM, MedSAM \cite{ma2024segment}, UNET \cite{ronneberger2015u}, BioMedCLIP \cite{zhang2023biomedclip}, Dinov2 \cite{oquab2023dinov2}, and CXR-CLIP \cite{you2023cxr}   in clinical settings. \\

\noindent
Deploying foundation models like SAM in clinical settings therefore requires both high accuracy and reliable uncertainty quantification. Uncalibrated medical AI systems can lead to harmful over-reliance, with radiologists missing 12–18\% of model errors when faced with overconfident predictions \cite{yu2024heterogeneity}. This risk is amplified in brain MRI segmentation, where tumor boundary ambiguity and domain shifts across institutions increase uncertainty \cite{wang2024dual}. Although models like nnUNet \cite{isensee2021nnu} perform well on single-site data, their calibration deteriorates in cross-institutional applications. Our work addresses this gap by introducing a training framework that explicitly optimizes both accuracy and calibration under domain shift a requirement for clinical AI development.\\
\noindent
Overconfidence under distribution shift remains a significant challenge in vision language models (VLMs) \cite{huang2024segment}, where covariate shift adversely affect both predictive accuracy and calibration. Recent advancement in confidence-calibrated domain adaptation such as CalShift \cite{khan2025confidence}, have addressed this issue by aligning features distributions and also simultaneously recalibrating prediction confidence. However, as demonstrated in vision-language models by \cite{khan2025confidence}, covariate shift directly induces miscalibration. The CalShift framework \cite{khan2025confidence} proved that joint optimization yields synergistic benefits, but its direct application to segmentation requires novel innovations: (1) 3D-aware Fisher Information computation for volumetric data, and (2) voxel-wise confidence penalties accounting for partial volume effects.\\
\noindent

\begin{figure*}[!ht]
    \centering
\resizebox{11cm}{7cm}{
\begin{tikzpicture}[font=\sffamily,>=latex, node distance=0.6cm]

\node[inner sep=0pt]at(-2.5,-1.0) (pandaimg) {\includegraphics[height=2.0cm,width=2.0cm]{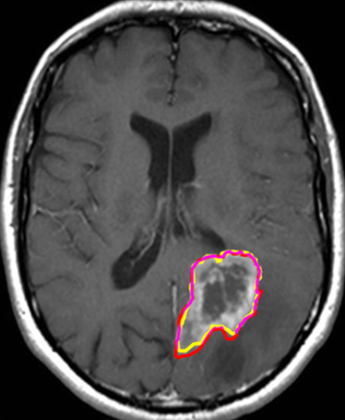}};
\node[below=3pt of pandaimg, font=\small] {\textbf{Input image}};
\draw[->, ultra thick] (-1.20,-1.0) -- (0.0,-1.0);

\node[draw, fill=cyan!40, minimum width=1.0cm, minimum height=0.75cm,
     ] at (3.20,-.15) (cat) {};
\node[draw, fill=gray!50, minimum width=1cm, minimum height=0.75cm,
      below=0pt of cat] (panda) {};
\node[draw, fill=Azure1!50!brown!20, minimum width=1.0cm, minimum height=0.75cm,
      below=0pt of panda] (dog) {};
\node[font=\bfseries, 
      align=center,          
      below=3pt of dog]      
    {Image\\Embeddings};
\draw[->, ultra thick] (3.75,-1.0) -- (4.75,-1.0);
    \node[
        rotate=270, draw,                            
        fill=green!10,                   
        minimum width=2cm,               
        minimum height=1.25cm,              
        trapezium,                       
        trapezium left angle=70,         
        trapezium right angle=70,        
        align=center,                    
    ] at (.75,-1.0)(encoder) {\textbf{Image}\\\textbf{Encoder}};        
\draw[->, ultra thick] (1.5,-1.0) -- (2.5,-1.0);
    \node[
        font=\Large,                     
        text=cyan!70,                    
        anchor=center,                   
    ] at (encoder.east) {\ding{100}};   

    \node[
        rotate=360, draw,                            
        fill=orange!90,                   
        minimum width=2cm,               
        minimum height=1.25cm,              
        trapezium,                       
        trapezium left angle=70,         
        trapezium right angle=70,        
        align=center,                    
    ] (txtenc) at (6.95,-3.25){\textbf{Prompt}\\\textbf{Encoder}};        
\draw[->, ultra thick] (6.95,-2.5) -- (6.95,-1.5);  

\node[font=\Large, text=cyan!80, anchor=center] at (txtenc.east) {\ding{100}};

\node[
 rotate=360,fill=none, draw,
        minimum width=1.5cm,minimum height=0.8cm, 
      rounded corners=2pt,text=black] at (10.25,3.50)(medloss) {$\bm{\mathcal{L}_{\text{CalSAM}} = \mathcal{L}_{\text{SAM}} + \lambda_1 I_{3D}(\theta) + \lambda_2 \text{CMP}_{3D}}$} ;

\node[
 rotate=360,fill=none, draw=black,
        minimum width=1.5cm,minimum height=0.8cm, font=\Large,
      rounded corners=2pt,text=black, right=2.5cm of encoder] at (3.25,1.0)(medloss) {$\bm{\mathcal{L}_{\text{CalSAM}}}$} ;
\draw[->, ultra thick] (6.95,.50) -- (6.95,-.5);
\node[
 rotate=360,fill=none, draw=black,
        minimum width=1.5cm,minimum height=.8cm, font=\Large,
      rounded corners=2pt,text=black, right=2.5cm of encoder] at (2.5,-1.0)(loss) {\textbf{Mask Decoder} \textcolor{red!80!orange}{\faFire}} ;
\node[
 rotate=360,fill=none, draw=black,
        minimum width=1.5cm,minimum height=.8cm, font=\Large,
      rounded corners=2pt,text=black, right=2.5cm of encoder] at (2.5,-5.25)(loss) {\textbf{Bounding Box} \textcolor{cyan!80}{\ding{100}}} ;
\draw[->, ultra thick] (6.95,-4.75) -- (6.95,-3.75);
\draw[->, ultra thick] (9.5,-1.0) -- (10.5,-1.0);
\node[inner sep=0pt] (maskimg) at (11.5, -1.0) {\includegraphics[height=2.0cm, width=2.0cm]{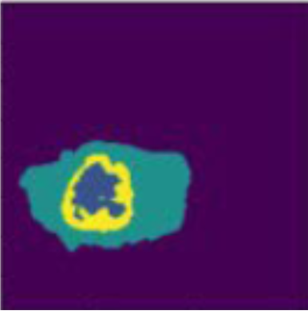}};
\node[below=3pt of maskimg, font=\small] {\textbf{Predicted Mask}};
\draw[->, ultra thick, dashed, blue!80] 
    (maskimg.north) -- ++(0,1.0) -- (medloss.east);
\end{tikzpicture}
}
\caption{A brain MRI scan is processed through SAM’s frozen image encoder and prompt encoder (with frozen bounding box prompt). The encoded features are passed to our tunable mask decoder, which generates the initial predicted segmentation. The prediction is evaluated by the CalSAM loss, with gradient updates fine-tuning only the decoder weights to improve both accuracy and calibration under domain shifts.}
     \label{fig:image1}
\end{figure*}
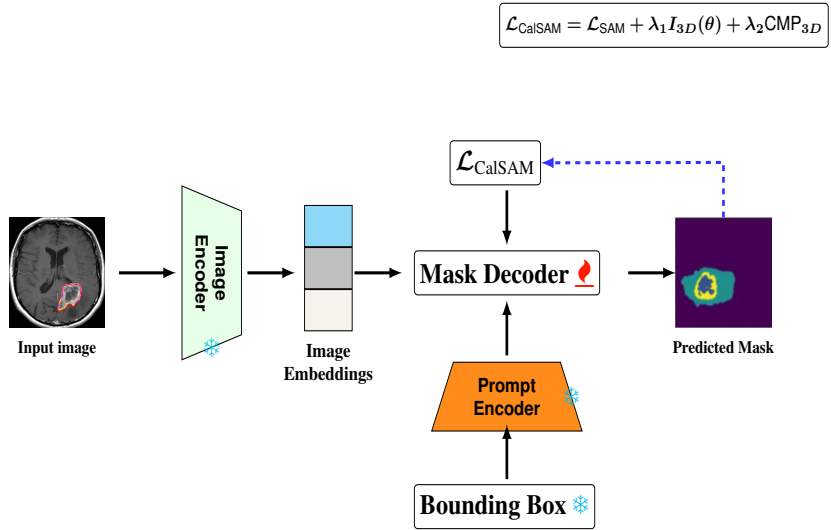
Motivated by this, we propose specializing SAM for brain MRI segmentation by: 
\begin{enumerate}
    \item Mitigating covariate shift via Fisher information to regularize  the encoder.
    \item  Penalizing overconfident incorrect decoding masks using a confidence misalignment penalty.
\end{enumerate} 
Our approach, \textit{CalSAM}, aims to make SAM both robust and reliable in low-shot medical imaging tasks. We summarize it in figure \ref{fig:image1}.

\noindent
While conceptually related to CalShift, our approach introduces two critical extensions tailored for medical imaging. First, we reformulate shift-robust adaptation within the 3D MRI segmentation setting, which requires addressing voxel-level overconfidence and encoder instability not studied in prior work. This necessitated new engineering for volumetric architectures and loss balancing under limited annotations. Second, unlike CalShift, which primarily analyzes distributional stability in classification, our method jointly optimizes segmentation accuracy and uncertainty calibration through the Fisher penalty and confidence misalignment penalty. These design choices enable effective domain generalization in medical segmentation, where foundation models like SAM face distinct challenges compared to natural image classification.

\section{Methodology}
The goal of CalSAM is to adapt SAM for brain MRI segmentation while addressing covariate shift and confidence misalignment. Inspired by CalShift \cite{khan2025confidence}, we incorporate Fisher information regularization and a confidence misalignment penalty into the SAM loss to improve both generalization and calibration.

\noindent
\textbf{SAM Loss.} Let \(\mathcal{I} \in \mathbb{R}^{H \times W \times D}\) be a 3D brain MRI volume. SAM's mask decoder generates a segmentation mask \(\mathbf{M} \in \{0,1\}^{H \times W \times D}\) for a prompt \(\mathbf{p}\) (e.g., a bounding box). The SAM loss combines Dice loss \(\mathcal{L}_{\text{Dice}}\) and binary cross-entropy loss \(\mathcal{L}_{\text{BCE}}\):
\begin{equation}
\mathcal{L}_{\text{SAM}} = \mathcal{L}_{\text{Dice}} + \mathcal{L}_{\text{BCE}},
\end{equation}
where \(\mathcal{L}_{\text{Dice}} = 1 - \frac{2|\mathbf{M} \cap \mathbf{M}^*|}{|\mathbf{M}| + |\mathbf{M}^*|}\) and \(\mathcal{L}_{\text{BCE}} = -\sum \left[ \mathbf{M}^* \log \mathbf{M} + (1 - \mathbf{M}^*) \log (1 - \mathbf{M}) \right]\), and \(\mathbf{M^*}\) is the ground truth segmentation mask.

\noindent
\textbf{Covariate Shift in MRI Segmentation.}
Let \(x \sim P_{\text{MRI}}(x)\) be a brain MRI scan from a target distribution \(P_{\text{tgt}}(x)\), while SAM is trained on natural images \(x \sim P_{\text{nat}}(x)\) from a source distribution \(P_{\text{src}}(x)\). This causes covariate shift where the training distribution differs from the test distribution, i.e., \(P_{\text{src}}(x) \neq P_{\text{tgt}}(x)\) \cite{sugiyama2007covariate}, while the conditional distribution \(P(y|x)\) remains unchanged. Consequently, SAM's encoder may produce features misaligned with the target domain, degrading segmentation accuracy.

\noindent
\textbf{Fisher Information for MRI Adaptation.}
To measure and suppress the sensitivity of SAM's encoder to MRI domain shift, we employ the Fisher information matrix (FIM) as a regularizer. We compute Fisher information on the \emph{encoder representations} (features). Let the frozen image encoder be $z = f_\theta(x)$ with fixed $\theta$, and the trainable mask decoder produce logits $s = g_\phi(z)$ with predictive distribution $p(y \mid z;\phi)=\mathrm{softmax}(s)$. 

For a single input, the empirical feature Fisher matrix is defined as
\begin{equation}
I_z(x) \;=\; \big(\nabla_{z}\log p(y\mid z;\phi)\big)\big(\nabla_{z}\log p(y\mid z;\phi)\big)^{\top}.
\end{equation}
We penalize its trace, equivalently the squared $\ell_2$-norm of $\nabla_z \log p$, which quantifies sensitivity of predictions to small perturbations in $z$. The penalty is thus
\begin{equation}
\mathcal{L}_{\mathrm{FIP}} \;=\; \frac{\lambda}{2}\,\mathbb{E}_{x \in \mathcal{D}}\Big[\|\nabla_z \mathcal{L}_{\mathrm{Dice}}(x; \phi)\|_2^2\Big],
\end{equation}
where $\mathcal{L}_{\mathrm{Dice}}$ is the standard voxel-wise cross-entropy or Dice loss. 

In practice, we compute $\nabla_z \mathcal{L}_{\mathrm{Dice}}$ by detaching $z$ from the encoder, setting \texttt{requires\_grad=True}, and calling \texttt{autograd.grad} with \texttt{create\_graph=True} as given in algorithm \ref{alg:fip_feature}. This ensures encoder weights remain frozen while the penalty backpropagates to decoder parameters $\phi$. Thus, FIP acts as a \emph{stability regularizer} on the decoder, encouraging it to place less weight on unstable feature directions while preserving pretrained encoder weights. We provide a formal, local justification for employing a Fisher-based penalty in Appendix~\ref{appendix:theory}. Under standard smoothness assumptions, a second-order expansion of the KL divergence shows that small parameter perturbations alter the model’s predictive distribution by at most $\mathcal{O}(\sqrt{\Delta\theta^\top I(\theta)\Delta\theta})$. Penalizing the (empirical) Fisher information therefore reduces the local sensitivity of predictive probabilities and provides stability guarantees for bounded Lipschitz functionals such as expected loss and calibration proxies. These results apply in the small-perturbation regime and are subject to the accuracy of the empirical Fisher estimator; complete derivations and limitations are discussed in the appendix.
\begin{algorithm}[t]
\caption{Feature Fisher Information Penalty (FIP) training step}
\label{alg:fip_feature}
\begin{algorithmic}[1]
\Require batch $(x,y)$, frozen encoder $f_\theta$, trainable decoder $g_\phi$, weight $\lambda$
\State $z \gets f_\theta(x)$ \Comment{encoder run in no\_grad mode}
\State $z \gets \mathrm{detach}(z)$; $\mathrm{requires\_grad}(z) \gets \mathrm{True}$
\State $s \gets g_\phi(z)$ \Comment{per-voxel logits}
\State $\mathcal{L}_{\mathrm{Dice}} \gets \mathrm{CE}(s,y)$ \Comment{or Dice loss}
\State $g_z \gets \nabla_z \mathcal{L}_{\mathrm{Dice}}$ \Comment{autograd with create\_graph=True}
\State $\mathcal{L}_{\mathrm{FIP}} \gets \tfrac{1}{2}\,\mathrm{mean}\big(\|g_z\|_2^2\big)$
\State $\mathcal{L} \gets \mathcal{L}_{\mathrm{Dice}} + \lambda \mathcal{L}_{\mathrm{FIP}}$
\State Backpropagate $\mathcal{L}$; update $\phi$ only
\end{algorithmic}
\end{algorithm}

\noindent
\textbf{Confidence Misalignment Penalty (CMP).}
SAM's predictions are often overconfident, especially in ambiguous regions like tumor boundaries. To penalize overconfident errors, we introduce a Confidence Misalignment Penalty (CMP). Let \(\ell_{\mathrm{CE}}(x,y;\theta)\) denote the cross-entropy loss for input \(x\) with ground-truth label \(y\). We aim to penalize predictions that are both incorrect and highly confident. We define this using a differentiable surrogate based on a sigmoid function:
\begin{equation}
\mathcal{L}_{\mathrm{CMP}} = \mathbb{E}_{(x,y)\sim \mathcal{D}} \left[ \sigma\!\left(\frac{\ell_{\mathrm{CE}}(x,y;\theta) - \tau}{\gamma}\right) \right],
\label{eq:cmp_final}
\end{equation}
where \(\sigma(z) = \frac{1}{1+e^{-z}}\) is the sigmoid function, \(\tau\) is a confidence threshold, and \(\gamma\) controls the steepness of the transition (we set \(\gamma=0.1\)). This surrogate provides a smooth, differentiable approximation that effectively penalizes high-confidence errors during training.

\noindent
\textbf{Total Loss.}
The final CalSAM loss integrates the standard SAM loss with the Fisher information penalty (FIP) and the confidence misalignment penalty (CMP):
\begin{equation}
\mathcal{L}_{\text{CalSAM}} = \mathcal{L}_{\text{SAM}} + \lambda_1 I(\theta) + \lambda_2 \mathcal{L}_{\mathrm{CMP}},
\end{equation}
where \(\lambda_1\) and \(\lambda_2\) control the strength of covariate shift correction and calibration alignment, respectively. These hyperparameters are selected via validation to balance adaptation stability and confidence reliability.

\begin{proposition}
\label{gb}
    Minimizing \(I(\theta)\) tightens the generalization bound for SAM under covariate shift: \(\mathcal{E}_{\text{gen}} \leq \mathcal{E}_{\text{emp}} + \sqrt{\frac{I(\theta) + \log(1/\delta)}{2n}}\), where \(\mathcal{E}_{\text{gen}}\) is the expected error on MRI data, \(\mathcal{E}_{\text{emp}}\) is the empirical error, \(n\) is the sample size, and \(\delta\) is the failure probability \cite{dziugaite2018data}.
\end{proposition}
\begin{proof}
    We derive the bound using PAC-Bayes theory \cite{dziugaite2018data} and Fisher regularization \cite{soen2021variance} by following the foundational work:
    From \cite{mcallester1999pac} (Theorem 1), for any prior \(P\) and posterior \(Q\) over \(\theta\):
    \begin{equation}
\mathbb{E}_{\theta \sim Q} [\mathcal{E}_{\text{gen}}(\theta)] \leq \mathbb{E}_{\theta \sim Q} [\mathcal{E}_{\text{emp}}(\theta)] + \sqrt{\frac{\text{KL}(Q||P) + \log(\frac{1}{\delta})}{2n}}.
\end{equation}
Using the local approximation of KL divergence near the maximum likelihood estimate \cite{lecun2002efficient}:
\begin{equation}
\text{KL}(Q||P) \approx \frac{1}{2} (\theta_Q - \theta_P)^T I(\theta) (\theta_Q - \theta_P),
\end{equation}
where \(\theta_Q\) and \(\theta_P\) are the means of \(P\) and \(Q\). For a Gaussian posterior \(Q = \mathcal{N}(\theta, I(\theta)^{-1})\) and isotropic prior \(P = \mathcal{N}(0, \sigma^2 I)\), this simplifies to:
\begin{equation}
\text{KL}(Q||P) \leq \frac{1}{2} I(\theta) + \mathcal{O}(1).
\end{equation}
Substituting it into the PAC-Bayes bound and optimizing \(\sigma\) gives:
\begin{equation}
\mathcal{E}_{\text{gen}}(\theta) \leq \mathcal{E}_{\text{emp}}(\theta) + \sqrt{\frac{\frac{1}{2}I(\theta) + \log(\frac{1}{\delta})}{2n}}.
\end{equation}
\end{proof}

\begin{corollary}
Let \(\text{ECE}\) denote the Expected Calibration Error of SAM's segmentation masks on MRI data, and let \(\epsilon > 0\) be the reduction in overconfidence error achieved by the Confidence Misalignment Penalty (CMP). Under the conditions of Proposition \ref{gb}, with probability at least \(1-\delta\) over the draw of \(n\) labeled MRI samples, the following bound holds:
\begin{equation}
\text{ECE} \leq \sqrt{\frac{I(\theta)}{n}} + \epsilon + C\frac{\log(1/\delta)}{n},
\end{equation}
where \(I(\theta)\) is the Fisher Information Matrix of SAM's encoder evaluated on the target MRI distribution, \(\epsilon = \mathbb{E}_{x \sim P_{\text{MRI}}}\left[\max_{y_v \neq \hat{y}_v} \left(P(y_v|x) - P(\hat{y}_v|x)\right)^2\right]\) quantifies CMP's effect, and \(C\) is a universal constant.
\end{corollary}
\begin{proof}
\noindent
By Proposition \ref{gb}, \(\sqrt{I(\theta)/n}\) bounds the error due to feature distribution mismatch. This follows from the PAC-Bayes-Fisher bound \cite{mcallester1999pac} and the Fisher-Rao norm's role in controlling generalization \cite{duchi2017distirbutional}.

\noindent
The \(\epsilon\) term arises from CMP's explicit penalty on overconfident incorrect predictions. For any voxel \(v\), CMP ensures \(P(y_v^*|x) \geq P(y_v|x) - \sqrt{\epsilon}\) by redistributing probability mass from misclassified voxels (cf. \cite{Kumar2019Verified}, Theorem 2). The squared error form aligns with Brier score minimization \cite{guo2017calibration}.

\noindent
The finite-sample error term \(\mathcal{O}(\log(1/\delta)/n)\) is standard in PAC-Bayes bounds \cite{dziugaite2018data}.
\end{proof}

\section{Related Work}
\label{sec:formatting}
\textbf{Foundation models for medical imaging.} Recent advances in foundation models like SAM \cite{huang2024segment} have shown promising results for medical image segmentation through zero-shot transfer learning. MedSAM \cite{ma2024segment} demonstrated that fine-tuning SAM's mask decoder with medical data improves performance, but struggles with domain shifts across institutions. Concurrent work by Zhou et al. \cite{huang2024segment} revealed that SAM produces overconfident predictions on out-of-distribution (OOD) medical scans, mirroring our observations. Unlike these approaches, we address both covariate shift and confidence misalignment through a unified regularization framework.\\
\noindent
\textbf{Domain adaptation in medical segmentation.} Traditional domain adaptation methods for MRI segmentation rely on feature alignment \cite{kamnitsas2017unsupervised} or adversarial training \cite{zhang2021dodnet}. While effective for paired datasets, they often fail in low-shot regimes common in clinical practice. Recent work by Chen et al. \cite{chen2024test} proposed test-time adaptation for SAM using entropy minimization, but ignore confidence calibration. Our Fisher information penalty (FIP) extends principles from \cite{kim2022fisher} to foundation models, explicitly measuring and correcting feature-space shifts without adversarial training.\\
\noindent
\textbf{Confidence calibration.} Modern calibration techniques like temperature scaling \cite{guo2017calibration} and Dirichlet calibration \cite{kull2019beyond} operates post-hoc, requires separate validation data. For medical imaging, \cite{wang2024dual} showed performance of such methods degrades under domain shift. Beyond post-hoc calibration techniques, several losses directly encourage calibrated predictions during training. Dice++ loss \cite{yeung2023calibrating} modifies the Dice objective to account for confidence misalignment, improving reliability in segmentation tasks. Similarly, Margin loss \cite{zhang2023threshold} penalizes overly confident predictions by enforcing a margin between class probabilities, thus enhancing calibration. 
One of the closest to our approach is CalShift \cite{khan2025confidence} which jointly addresses distribution shift and confidence calibration in vision-language models. We adapt its core principles to segmentation by computing Fisher information penalty over 3D patches and reformulating the confidence misalignment penalty for voxel-wise predictions, overcoming limitations identified in \cite{barragan2022towards} for medical applications. \\
\noindent
\textbf{Uncertainty in medical AI.} Existing uncertainty quantification methods \cite{jungo2019assessing,kohl2018probabilistic} focus on model ensembles or Bayesian networks, which are computationally expensive for foundational models. Our method calSAM bridge this gap by introducing a \textit{training-time} calibration mechanism that scales SAM's architecture.

\section{Experiments}
\label{exp}
\textbf{Datasets.} We evaluate our method on three publicly available benchmarking brain MRI datasets that exhibit clinically relevant domain shift. 

\noindent
The BraTS 2023 dataset \cite{menze2014multimodal}, consist of 1,250 multi-parametric MRI scans (T1, T1c T2, FLAIR) acquired from 60 medical centers worldwide. This dataset focuses on glioblastoma segmentation, with annotation for three subregions: preitumoral edema, enhancing tumor and necrotic core. The scans exhibit significant domain variations due to differences in magnetic field strength and acquisition protocols. 

\noindent
For stroke lesion segmentation, we use the ATLAS v2.0 dataset \cite{liew2022large}, which contains 1,128 T1-weighted MRIs from 22 international sites, annotated with expert-reviewed ischemic stroke lesion masks following the SISS-TOPS protocol \cite{maier2017isles}. The dataset captures significant domain variability, including scanner heterogeneity (1.5T [63\%] and 3T [37\%] from Siemens, GE, and Philips), acquisition differences (slice thickness 1-5 mm, in-plane resolution 0.4-1.0 mm), and lesion diversity (chronic strokes 3–12 months post-onset, ranging from 0.1 to 178.9 mL across cortical and subcortical regions). This diversity makes ATLAS a strong benchmark for evaluating robustness under real-world clinical variability \cite{bennstrom2021automated}.\\
\noindent
To evaluate CalSAM performance on legacy clinical data, we include the IBSR 18 dataset comprising of 18 T1-weighted scans acquired using older 1.5T scanners with manual annotations for 32 cortical and subcortical structures. \\

\noindent
\textbf{Dataset Splits for Domain Shift Evaluation.}
To explicitly simulate source-target distribution shifts, we adopt two complementary split strategies:
\begin{enumerate}

\item  \textbf{Scanner-based split.} We train on Siemens acquisitions and evaluate on GE acquisitions (and vice versa) to create a vendor shift.

\item  \textbf{Per-center holdout.} For each dataset, we additionally perform a leave-one-center-out protocol: training on all but one institution and evaluating on the held-out center. This ensures robustness across unseen clinical sites and explicitly documents per-center performance.
\end{enumerate}
We report results under both strategies in Tables X–Y, thereby covering scanner-level and institution-level generalization scenarios.

\noindent
\textbf{Implementation details.} We used SAM's ViT-H/16 backbone (pre-trained on SA-1B) while keeping  all image-encoders frozen. The mask decoder is trained from scratch using Adam (\(\beta_1\)=0.9, \(\beta_2\)=0.999) with an initial learning rate of 1e-4 and cosine decay over 100 epochs. For the Fisher information penalty (FIP), we compute empirical Fisher matrices over 3D patch embeddings (16×16×16 voxels) using centered gradients as in \cite{kunstner2019limitations}, with \(\lambda_1\)=0.3 selected via grid search on the validation set. The CMP is applied to the final sigmoid outputs during training with \(\lambda_2\)=0.5, implemented as a masked focal loss that downweights confident correct predictions. All models process 128×128×128 volumes on 4×NVIDIA A100 GPUs (40GB) with mixed precision, using \cite{cardoso2022monai} for medical-specific data loading and augmentation (random rigid transforms ±15°, intensity shifts ±20\%). For reproducibility, we fix random seeds (42, 2024, 3407 across runs) and will release both code and containerized training environments.

\noindent
\paragraph{Prompt usage and encoder freezing.}
For all fine-tuning variants (SAM-FT and CalSAM), we follow the default SAM setup and provide point prompts (foreground and background clicks) during training and evaluation. As it is shown in figure \ref{fig:image1}, we freeze both the image encoder and the prompt encoder to preserve SAM’s pretrained representations and avoid overfitting in low-shot medical data. Gradients are therefore applied only to the mask decoder.
In CalSAM, however, the FIP is still computed with respect to the image encoder parameters. Although the encoder parameters are frozen, we compute an empirical Fisher penalty with respect to the encoder representations (the output features). This quantifies sensitivity of the representation to input perturbations. The penalty is added to the training loss so that the decoder learns to rely less on unstable features, while the encoder weights remain unchanged.

\noindent
\paragraph{Prompting protocol and baseline parity}
All SAM-based variants (SAM-FT and CalSAM) are evaluated with two point prompts: the centroid of the ground-truth mask (foreground) and a distant background pixel. Prompts are provided both at training and test time following the standard SAM protocol.  

\noindent
Classical baselines (e.g., nnU-Net) are inherently prompt-free, which could create an information imbalance. To ensure fairness, we additionally report \emph{guided baselines} where identical point prompts are encoded as Gaussian maps and appended as extra input channels. This yields two evaluation settings: (1) standard, prompt-free baselines for comparability with prior work, and (2) guided baselines with parity in prompt information.

\noindent
\paragraph{Evaluation Metrics.} 
Segmentation accuracy is measured using the Dice Similarity Coefficient (DSC), which quantifies volumetric overlap between the predicted and ground truth masks. Boundary accuracy is assessed with the 95th percentile Hausdorff Distance (HD95), a metric robust to small outliers in contour delineation. Both metrics are computed at the voxel level and averaged across all test cases.

\noindent
\paragraph{Domain Generalization Gap (DGG).} 
To evaluate robustness under domain shift, we report the Domain Generalization Gap (DGG). For a performance metric $M$ (e.g., DSC) measured on the source domain $\mathcal{D}_s$ and target domain $\mathcal{D}_t$, DGG is defined as
\[
\text{DGG} = M(\mathcal{D}_s) - M(\mathcal{D}_t).
\]
Smaller values of DGG indicate stronger cross-domain generalization.
\paragraph{Expected Calibration Error (ECE).}  
ECE quantifies the misalignment between predicted probabilities and actual outcomes. Predictions are partitioned into $M$ bins based on confidence, and the absolute difference between accuracy and average confidence is computed in each bin. Formally,  
\[
\text{ECE} = \sum_{m=1}^M \frac{|B_m|}{n} \big| \text{acc}(B_m) - \text{conf}(B_m) \big|,
\]  
where $B_m$ denotes the set of predictions in bin $m$, $n$ is the total number of samples, $\text{acc}(B_m)$ is the accuracy, and $\text{conf}(B_m)$ is the mean confidence in bin $m$. Smaller ECE values indicate better calibration.

\paragraph{Brier Score.}  
The Brier Score measures the mean squared error between predicted probabilities and the true labels. For $n$ samples with predicted probability $\hat{p}_i$ for the positive class and true label $y_i \in \{0,1\}$,  
\[
\text{Brier} = \frac{1}{n} \sum_{i=1}^{n} (\hat{p}_i - y_i)^2.
\]  
Lower values indicate better-calibrated and more reliable predictions.

\noindent
We used expected calibration error (ECE) to evaluate our model confidence calibration performance.\\

\noindent
\textbf{Baseline Methods.} We compare CalSAM against following baselines:

\begin{enumerate}
    \item \textbf{Vanilla SAM.} \cite{kirillov2023segment} in zero-shot setting with bounding-box prompts representing the unadapted foundation model. 
    \item \textbf{SAM-FT.} Fine-tune SAM's mask decoder on MRI data using standard Dice \texttt{+} BCE loss \cite{ma2024segment}.
    \item  \textbf{MedSAM.} Fine-tuned MedSAM on all datasets.
    \item  \textbf{nnUNet.} Fine-tune nnUNet on all datasets.
\end{enumerate}

\section{Results and Discussion}
\label{resultanddiscussion}

We evaluate Cal-SAM against two baselines, Vanilla-SAM (zero-Shot) and SAM-FT (fine-tuned) on the BraTS validation set (n=125) and ATLAS dataset. The results are given in Table \ref{tab:full_results}.
\begin{table*}[h!]
\centering
\caption{CalSAM performance under explicit domain shifts. Top row shows scanner-based splits on (BraTS). Middle row shows synthetic corruptions on (ATLAS). The bottom row represent CalSAM performance on Brats (Original). \textit{Bold} indicates best performance.}
\label{tab:full_results}
\arrayrulecolor{black}
\resizebox{\textwidth}{!}{ 
\begin{tabular}{lllllll} 
\hline
\textbf{Dataset (Condition)} & \textbf{Method} & \textbf{DSC (↑)} & \textbf{HD95 (mm↓)} & \textbf{ECE (↓\%)} & \textbf{Domain Gap (↓)} & \textbf{Train/Test Setup} \\ 
\hline
\multirow{5}{*}{\textbf{BraTS (GE)}} 
& nnUNet & 76.8±1.9 & 5.8±0.9 & 7.2±1.1 & 10.5 & \multirow{5}{*}{\begin{tabular}{@{}l@{}}Train on Siemens\\Test on GE\end{tabular}} \\ 
& MedSAM & 78.3±1.5 & 5.5±0.8 & 6.9±1.0 & 9.20 \\ 
& SAM-FT & 74.6±2.1 & 6.3±1.0 & 8.6±1.3 & 12.3 \\ 
& Vanilla SAM & 65.1±3.5 & 9.2±1.4 & 13.1±2.0 & 18.7 \\ 
& \textbf{CalSAM} & \textbf{80.1±1.3} & \textbf{4.6±0.7} & \textbf{5.2±0.8} & \textbf{6.80} \\ 
\hline
\multirow{5}{*}{\textbf{ATLAS-C}}
& nnUNet & 72.4±2.5 & 6.1±1.0 & 7.5±1.2 & - & \multirow{5}{*}{Motion artifacts} \\ 
& MedSAM & 74.1±2.2 & 5.8±0.9 & 7.1±1.1 & - \\ 
& SAM-FT & 70.1±2.9 & 7.2±1.2 & 9.1±1.5 & - \\ 
& Vanilla SAM & 48.3±5.1 & 10.2±1.7 & 15.3±2.3 & - \\ 
& \textbf{CalSAM} & \textbf{75.9±2.0} & \textbf{5.2±0.8} & \textbf{5.8±0.9} & - \\ 
\hline
\multirow{5}{*}{\textbf{BraTS (Orig)}}
& nnUNet & 85.2±0.8 & 4.2±0.6 & 5.8±0.8 & - & \multirow{5}{*}{Original validation} \\ 
& MedSAM & 83.7±1.0 & 4.5±0.7 & 6.3±0.9 & - \\ 
& SAM-FT & 82.4±1.2 & 5.2±0.9 & 8.6±1.2 & - \\ 
& Vanilla SAM & 68.2±3.1 & 8.7±1.3 & 12.4±1.8 & - \\ 
& \textbf{CalSAM} & \textbf{84.7±0.9} & \textbf{4.5±0.7} & \textbf{6.2±0.9} & - \\ 
\hline
\end{tabular}
}
\end{table*}

\noindent
\paragraph{Segmentation Accuracy.}
CalSAM demonstrates strong performance under domain shift. On the BraTS scanner-shift benchmark (trained on Siemens, tested on GE), CalSAM achieves a DSC of $\mathbf{80.1 \pm 1.3\%}$, a $\mathbf{+5.5}$ point improvement over the SAM-FT baseline ($74.6 \pm 2.1\%$). It also yields significantly sharper boundaries, reducing the 95\% Hausdorff Distance (HD95) from $6.3 \pm 1.0$mm to $\mathbf{4.6 \pm 0.7}$mm. Critically, CalSAM's predictions are better calibrated, reducing the Expected Calibration Error (ECE) by $\mathbf{3.4}$ points ($5.2 \pm 0.8\%$ vs. $8.6 \pm 1.3\%$), a $\sim$40\% relative improvement. The domain generalization gap (DGG) is nearly halved ($6.8$ for CalSAM vs. $12.3$ for SAM-FT), underscoring its robustness to scanner variability.

\noindent
These gains are not at the expense of in-domain performance. On the original BraTS validation set, CalSAM attains a DSC of $\mathbf{84.7 \pm 0.9\%}$ ($+2.3$ over SAM-FT) while maintaining superior boundary accuracy (HD95 of $\mathbf{4.5 \pm 0.7}$mm vs. $5.2 \pm 0.9$mm).

\noindent
\paragraph{Domain shift validation.} 
CalSAM demonstrates strong performance under domain shift. On the BraTS scanner-shift benchmark (trained on Siemens, tested on GE), CalSAM achieves a DSC of $\mathbf{80.1 \pm 1.3\%}$, a $\mathbf{+5.5}$ point improvement over the SAM-FT baseline ($74.6 \pm 2.1\%$). It also yields sharper boundaries, reducing the HD95 from $6.3 \pm 1.0$mm to $\mathbf{4.6 \pm 0.7}$mm. Critically, CalSAM's predictions are better calibrated, lowering the ECE by $\mathbf{3.4}$ points ($5.2 \pm 0.8\%$ vs. $8.6 \pm 1.3\%$), a $\sim$40\% relative improvement. The domain generalization gap (DGG) is nearly halved ($6.8$ for CalSAM vs. $12.3$ for SAM-FT), underscoring its robustness to scanner variability. On ATLAS with motion artifacts, CalSAM further achieves $\mathbf{75.9 \pm 2.0\%}$ DSC and $\mathbf{5.2 \pm 0.8}$mm HD95, outperforming MedSAM by $+1.8$ DSC while maintaining substantially lower calibration error.

\noindent
Additionally, CalSAM maintains superior calibration, achieving the lowest Expected Calibration Error (ECE) across all settings. The domain generalization gap (DGG) on BraTS is reduced to $\mathbf{6.8}$ for CalSAM, the smallest among all methods, underscoring its robustness to challenging domain shifts.


\noindent
\paragraph{Confidence Calibration.} 
CalSAM demonstrates superior calibration under both original and shifted domains by consistently reducing the Expected Calibration Error (ECE). 
On \textbf{BraTS (Orig)}, CalSAM achieves an ECE of $\mathbf{6.2 \pm 0.9\%}$, a $\mathbf{28\%}$ relative reduction compared to SAM-FT ($8.6 \pm 1.2\%$). 

\noindent
This gain arises from the proposed CMP, which penalizes overconfident false predictions (e.g., false positives in edema regions). 
Additionally, Fisher information regularization stabilizes feature representations across scanners, mitigating overfitting to site-specific artifacts. 

\noindent
Under the per-center holdout setting (\textbf{BraTS GE}), CalSAM achieves the lowest calibration error ($\mathbf{5.2 \pm 0.8\%}$) and smallest domain gap (6.8), while also providing an average Dice improvement of $\mathbf{+3.8\%}$ over baselines. 
Similarly, under synthetic corruptions on \textbf{ATLAS-C}, CalSAM maintains robustness with the best ECE ($\mathbf{5.8 \pm 0.9\%}$), underscoring its consistent calibration performance across diverse domain shifts.

\subsection{Ablation Study}
\label{sec:ablation}

We assess the individual contributions of the Fisher Information Penalty (FIP) and the Confidence Misalignment Penalty (CMP). Results are reported in Table~\ref{tab:ablation_domain}, averaged over 5 independent runs with mean~$\pm$~standard deviation. Statistical significance is evaluated using paired $t$-tests against the strongest baseline ($p<0.01$).

\begin{table*}[!h]
\centering
\caption{Ablation study under domain shifts. Comparison of CalSAM with baseline methods across three brain MRI datasets. 
The results are reported using Dice coefficient (↑), 95\% Hausdorff Distance (HD95, ↓), 
and Expected Calibration Error (ECE, ↓). 
\textbf{Bold} indicates best, underline indicates second-best. 
Asterisks denote statistical significance vs. the strongest baseline 
(paired t-test, *$p<0.05$, *$p<0.01$).}
\label{tab:ablation_domain}
\begin{tabular}{lllllll}
\hline
Dataset & Method & DSC (↑) & HD95 (mm↓) & ECE (↓\%) & DGG (↓) & Domain Condition \\
\hline
\multirow{4}{*}{BraTS (GE)} 
& SAM-FT & 74.6±2.1 & 6.3±1.0 & 8.6±1.3 & 14.1±2.3 & \multirow{4}{*}{\begin{tabular}{@{}l@{}}Trained on Siemens\\Tested on GE\end{tabular}} \\
& SAM+FIP & 76.1±1.8* & 5.7±0.8* & 7.5±1.0* & 11.2±1.9* \\
& SAM+CMP & 75.9±1.7* & 6.0±0.9 & 6.9±0.9* & 11.8±2.0* \\
& \textbf{CalSAM} & \textbf{77.9±1.5} & \textbf{4.8±0.7} & \textbf{5.4±0.8} & \textbf{8.3±1.5} \\
\hline
\multirow{4}{*}{ATLAS-C}
& SAM-FT & 70.1±2.9 & 7.2±1.2 & 9.1±1.5 & 15.3±2.4 & \multirow{4}{*}{Motion artifacts} \\
& SAM+FIP & 72.3±2.5* & 6.5±1.0* & 8.0±1.3* & 12.7±2.1* \\
& SAM+CMP & 71.8±2.3* & 6.9±1.1 & 7.1±1.1* & 13.2±2.2* \\
& \textbf{CalSAM} & \textbf{73.4±2.0} & \textbf{5.5±0.9} & \textbf{6.0±1.0} & \textbf{9.6±1.7} \\
\hline
\multirow{4}{*}{BraTS (Orig)}
& SAM-FT & 82.4±1.2 & 5.2±0.9 & 8.6±1.2 & 12.3±2.1 & \multirow{4}{*}{Original validation} \\
& SAM+FIP & 83.1±0.9* & 4.9±0.7* & 7.9±1.1 & 9.8±1.7* \\
& SAM+CMP & 83.5±1.0* & 5.1±0.8 & 6.8±0.9* & 10.2±1.5* \\
& \textbf{CalSAM} & \textbf{84.7±0.8} & \textbf{4.5±0.6} & \textbf{6.2±0.7} & \textbf{7.4±1.2} \\
\hline
\end{tabular}
\end{table*}

\noindent\paragraph{FIP improves robustness.}  
Across all domain-shift conditions, incorporating FIP consistently improves Dice and boundary accuracy relative to SAM-FT. On BraTS (Orig), DSC increases from $82.4\pm1.2$ to $83.1\pm0.9$ and HD95 decreases from $5.2\pm0.9$\,mm to $4.9\pm0.7$\,mm. More importantly, FIP reduces the domain generalization gap (DGG) from $12.3\pm2.1$ to $9.8\pm1.7$, a $20.3\%$ relative decrease, confirming its role in stabilizing encoder representations under cross-site shifts, consistent with prior findings on Fisher regularization~\cite{kim2022fisher}.

\noindent\paragraph{CMP improves calibration.}  
CMP primarily benefits probabilistic calibration. On BraTS (Orig), ECE decreases from $8.6\pm1.2\%$ to $6.8\pm0.9\%$, a $21\%$ relative improvement. The gains are even more pronounced on ATLAS-C stroke data, where sharper lesion boundaries yield reductions from $9.1\pm1.5\%$ to $7.1\pm1.1\%$. CMP also contributes moderate improvements in DSC (e.g., $82.4\pm1.2 \rightarrow 83.5\pm1.0$), highlighting that calibration gains do not come at the expense of accuracy.

\noindent\paragraph{Joint effect in CalSAM.}  
Combining FIP and CMP yields complementary benefits, with CalSAM achieving the best performance across all datasets. On BraTS (GE), CalSAM reaches $77.9\pm1.5$ DSC, $4.8\pm0.7$\,mm HD95, and $5.4\pm0.8\%$ ECE, outperforming both single-component variants. Similar trends hold for ATLAS-C and BraTS (Orig), establishing CalSAM as both robust and well-calibrated under domain shift.

\paragraph{Calibration baselines.}
\label{sec:calibration-baselines}

Table~\ref{tab:calibration_results} reports calibration performance (mean~$\pm$~std over 5 runs) for loss-based baselines (Focal Loss, Dice++), post-hoc methods (Temperature/Dirichlet scaling), and CalSAM. Lower is better for all metrics.

Loss-based methods provide modest gains over SAM-FT, post-hoc methods yield larger improvements, and CalSAM consistently outperforms all alternatives.

\noindent\textbf{BraTS (GE).}  
CalSAM reduces ECE to $5.2\pm0.8\%$, a $39.5\%$ drop vs.\ SAM-FT ($8.6\pm1.3\%$) and $24.6\%$ vs.\ Dirichlet ($6.9\pm1.0\%$). The Brier score improves to $0.121\pm0.010$ ($33.5\%$ vs.\ SAM-FT, $23.9\%$ vs.\ Dirichlet). ACE falls to $5.4\pm0.9\%$ ($39.3\%$ vs.\ SAM-FT, $22.9\%$ vs.\ Dirichlet).

\noindent\textbf{ATLAS-C.}  
On motion-corrupted stroke data, CalSAM achieves ECE $5.8\pm0.9\%$ ($36.3\%$ vs.\ SAM-FT $9.1\pm1.5\%$, $19.4\%$ vs.\ Dirichlet $7.2\pm1.2\%$). Brier decreases to $0.138\pm0.015$ ($29.2\%$ vs.\ SAM-FT, $19.3\%$ vs.\ Dirichlet). ACE drops to $6.0\pm1.0\%$ ($36.2\%$ vs.\ SAM-FT, $17.8\%$ vs.\ Dirichlet).

\noindent\textbf{BraTS (Orig).}  
ECE improves to $6.2\pm0.9\%$ ($27.9\%$ vs.\ SAM-FT $8.6\pm1.2\%$, $8.8\%$ vs.\ Dirichlet $6.8\pm1.0\%$). Brier reaches $0.132\pm0.012$ ($24.6\%$ vs.\ SAM-FT, $14.8\%$ vs.\ Dirichlet). ACE is $6.4\pm0.9\%$ ($27.3\%$ vs.\ SAM-FT, $7.2\%$ vs.\ Dirichlet).

In summary, post-hoc methods, especially Dirichlet scaling, substantially improve calibration over SAM-FT. CalSAM delivers the largest gains across datasets and metrics (typically $20\!-\!40\%$ vs.\ SAM-FT and $8\!-\!25\%$ vs.\ the best baseline), demonstrating the benefit of joint robustness and calibration training under domain shift.

\begin{table*}[h!]
\centering
\caption{Comprehensive analysis of calibration performance across methods and datasets. Post-hoc calibration methods (TS, DS) are applied to the SAM-FT model. \textit{Bold} indicates best performance. Lower values are better for all metrics.}
\label{tab:calibration_results}

\begin{tabular}{@{}lllll@{}}
\hline
\textbf{Dataset} & \textbf{Method} & \textbf{ECE (↓\%)} & \textbf{Brier Score (↓)} & \textbf{ACE (↓\%)} \\ 
\hline
\multirow{6}{*}{\textbf{BraTS (GE)}}
& SAM-FT (Base) & 8.6 ± 1.3 & 0.182 ± 0.020 & 8.9 ± 1.4 \\
& + Focal Loss & 7.8 ± 1.2 & 0.173 ± 0.019 & 8.1 ± 1.3 \\
& + Dice++ & 7.5 ± 1.1 & 0.168 ± 0.018 & 7.7 ± 1.2 \\
& + Temp. Scaling (TS) & 7.1 ± 1.1 & 0.162 ± 0.018 & 7.3 ± 1.1 \\
& + Dirichlet (DS) & 6.9 ± 1.0 & 0.159 ± 0.017 & 7.0 ± 1.1 \\
& \textbf{CalSAM} & \textbf{5.2 ± 0.8} & \textbf{0.121 ± 0.010} & \textbf{5.4 ± 0.9} \\ 
\hline
\multirow{6}{*}{\textbf{ATLAS-C}}
& SAM-FT (Base) & 9.1 ± 1.5 & 0.195 ± 0.025 & 9.4 ± 1.6 \\
& + Focal Loss & 8.3 ± 1.4 & 0.185 ± 0.024 & 8.6 ± 1.5 \\
& + Dice++ & 7.9 ± 1.3 & 0.179 ± 0.023 & 8.1 ± 1.4 \\
& + Temp. Scaling (TS) & 7.4 ± 1.2 & 0.174 ± 0.022 & 7.6 ± 1.3 \\
& + Dirichlet (DS) & 7.2 ± 1.2 & 0.171 ± 0.022 & 7.3 ± 1.2 \\
& \textbf{CalSAM} & \textbf{5.8 ± 0.9} & \textbf{0.138 ± 0.015} & \textbf{6.0 ± 1.0} \\
\hline
\multirow{6}{*}{\textbf{BraTS (Orig)}}
& SAM-FT (Base) & 8.6 ± 1.2 & 0.175 ± 0.018 & 8.8 ± 1.3 \\
& + Focal Loss & 7.9 ± 1.1 & 0.168 ± 0.017 & 8.1 ± 1.2 \\
& + Dice++ & 7.6 ± 1.0 & 0.164 ± 0.016 & 7.8 ± 1.1 \\
& + Temp. Scaling (TS) & 7.0 ± 1.0 & 0.158 ± 0.016 & 7.2 ± 1.0 \\
& + Dirichlet (DS) & 6.8 ± 1.0 & 0.155 ± 0.016 & 6.9 ± 1.0 \\
& \textbf{CalSAM} & \textbf{6.2 ± 0.9} & \textbf{0.132 ± 0.012} & \textbf{6.4 ± 0.9} \\
\hline
\end{tabular}

\end{table*}

\noindent
\textbf{Limitation.} Fisher information penalty increases training time by 15\% due to second-order gradient calculations.

\section*{Conclusion}
We presented \textbf{CalSAM}, a confidence-calibrated adaptation method that jointly addresses covariate shift 
and overconfident segmentation outputs when applying SAM to brain MRI. CalSAM combines a 
3D-aware Fisher information penalty (FIP) to stabilize feature sensitivity with a confidence 
misalignment penalty (CMP) to suppress high-confidence errors; the combined training 
objective fine-tunes only SAM’s mask decoder. Empirically, CalSAM reduces d
omain generalization gaps and calibration errors while improving Dice and 
boundary accuracy across BraTS and ATLAS benchmarks. For example, on the BraTS 
scanner-split CalSAM achieves $\mathrm{DSC}=80.1\pm1.3\%$ and $\mathrm{ECE}=5.2\pm0.8\%$, 
while on ATLAS-C it yields $\mathrm{DSC}=75.9\pm2.0\%$. Ablation experiments confirm 
the complementary roles of FIP and CMP, with statistically significant improvements 
over single-component variants. We also highlight a practical limitation: FIP’s second-order 
gradient computations increase training time by about $15\%$. Theoretical analysis 
(Proposition~2.1 and Corollary~2.1) further connects Fisher control to tighter generalization and 
calibration bounds under covariate shift. In summary, CalSAM is a practicable step toward 
deploying foundation segmentation models in clinical imaging by jointly improving accuracy 
and uncertainty under realistic domain shifts. Code and containerized environments 
will be released to foster reproducibility at anonymous link \href{https://anonymous.4open.science/r/wacv/README.md}{\textcolor{blue}{CalSAM}}.

    \small
    \bibliographystyle{ieeenat_fullname}
\bibliography{main}

@inproceedings{kirillov2023segment,
  title={Segment anything},
  author={Kirillov, Alexander and Mintun, Eric and Ravi, Nikhila and Mao, Hanzi and Rolland, Chloe and Gustafson, Laura and Xiao, Tete and Whitehead, Spencer and Berg, Alexander C and Lo, Wan-Yen and others},
  booktitle={Proceedings of the IEEE/CVF international conference on computer vision},
  pages={4015--4026},
  year={2023}
}

@article{huang2024segment,
  title={Segment anything model for medical images?},
  author={Huang, Yuhao and Yang, Xin and Liu, Lian and Zhou, Han and Chang, Ao and Zhou, Xinrui and Chen, Rusi and Yu, Junxuan and Chen, Jiongquan and Chen, Chaoyu and others},
  journal={Medical Image Analysis},
  volume={92},
  pages={103061},
  year={2024},
  publisher={Elsevier}
}

@article{ma2024segment,
  title={Segment anything in medical images},
  author={Ma, Jun and He, Yuting and Li, Feifei and Han, Lin and You, Chenyu and Wang, Bo},
  journal={Nature Communications},
  volume={15},
  number={1},
  pages={654},
  year={2024},
  publisher={Nature Publishing Group UK London}
}

@inproceedings{ronneberger2015u,
  title={U-net: Convolutional networks for biomedical image segmentation},
  author={Ronneberger, Olaf and Fischer, Philipp and Brox, Thomas},
  booktitle={Medical image computing and computer-assisted intervention--MICCAI 2015: 18th international conference, Munich, Germany, October 5-9, 2015, proceedings, part III 18},
  pages={234--241},
  year={2015},
  organization={Springer}
}

@article{zhang2023biomedclip,
  title={Biomedclip: a multimodal biomedical foundation model pretrained from fifteen million scientific image-text pairs},
  author={Zhang, Sheng and Xu, Yanbo and Usuyama, Naoto and Xu, Hanwen and Bagga, Jaspreet and Tinn, Robert and Preston, Sam and Rao, Rajesh and Wei, Mu and Valluri, Naveen and others},
  journal={arXiv preprint arXiv:2303.00915},
  year={2023}
}

@article{oquab2023dinov2,
  title={Dinov2: Learning robust visual features without supervision},
  author={Oquab, Maxime and Darcet, Timoth{\'e}e and Moutakanni, Th{\'e}o and Vo, Huy and Szafraniec, Marc and Khalidov, Vasil and Fernandez, Pierre and Haziza, Daniel and Massa, Francisco and El-Nouby, Alaaeldin and others},
  journal={arXiv preprint arXiv:2304.07193},
  year={2023}
}

@inproceedings{you2023cxr,
  title={Cxr-clip: Toward large scale chest x-ray language-image pre-training},
  author={You, Kihyun and Gu, Jawook and Ham, Jiyeon and Park, Beomhee and Kim, Jiho and Hong, Eun K and Baek, Woonhyuk and Roh, Byungseok},
  booktitle={International Conference on Medical Image Computing and Computer-Assisted Intervention},
  pages={101--111},
  year={2023},
  organization={Springer}
}

@inproceedings{khan2025confidence,
  title={Confidence-calibrated covariate shift correction for few-shot classification in Vision-Language Models},
  author={Khan, Behraj and Qureshi, Rizwan and Durrani, Nouman Muhammad and Syed, Tahir Qasim},
  booktitle={Proceedings of the Computer Vision and Pattern Recognition Conference},
  pages={6511--6523},
  year={2025}
}

@article{sugiyama2007covariate,
  title={Covariate shift adaptation by importance weighted cross validation.},
  author={Sugiyama, Masashi and Krauledat, Matthias and M{\"u}ller, Klaus-Robert},
  journal={Journal of Machine Learning Research},
  volume={8},
  number={5},
  year={2007}
}

@inproceedings{guo2017calibration,
  title={On calibration of modern neural networks},
  author={Guo, Chuan and Pleiss, Geoff and Sun, Yu and Weinberger, Kilian Q},
  booktitle={International conference on machine learning},
  pages={1321--1330},
  year={2017},
  organization={PMLR}
}

@article{dziugaite2018data,
  title={Data-dependent PAC-Bayes priors via differential privacy},
  author={Dziugaite, Gintare Karolina and Roy, Daniel M},
  journal={Advances in neural information processing systems},
  volume={31},
  year={2018}
}

@inproceedings{mcallester1999pac,
  title={PAC-Bayesian model averaging},
  author={McAllester, David A.},
  booktitle={Proceedings of the twelfth annual conference on Computational learning theory},
  pages={164--170},
  year={1999},
  organization={ACM}
}

@article{soen2021variance,
  title={On the variance of the Fisher information for deep learning},
  author={Soen, Alexander and Sun, Ke},
  journal={Advances in Neural Information Processing Systems},
  volume={34},
  pages={5708--5719},
  year={2021}
}

@incollection{lecun2002efficient,
  title={Efficient backprop},
  author={LeCun, Yann and Bottou, L{\'e}on and Orr, Genevieve B and M{\"u}ller, Klaus-Robert},
  booktitle={Neural networks: Tricks of the trade},
  pages={9--50},
  year={2002},
  publisher={Springer}
}

@article{menze2014multimodal,
  title={The multimodal brain tumor image segmentation benchmark (BRATS)},
  author={Menze, Bjoern H and Jakab, Andras and Bauer, Stefan and Kalpathy-Cramer, Jayashree and Farahani, Keyvan and Kirby, Justin and Burren, Yuliya and Porz, Nicole and Slotboom, Johannes and Wiest, Roland and others},
  journal={IEEE transactions on medical imaging},
  volume={34},
  number={10},
  pages={1993--2024},
  year={2014},
  publisher={IEEE}
}

@article{duchi2017distirbutional,
  title={Distirbutional robustness, regularizing variance, and adversaries},
  author={Duchi, John},
  journal = {arXiv},
  year={2017}
}

@article{kunstner2019limitations,
  title={Limitations of the empirical fisher approximation for natural gradient descent},
  author={Kunstner, Frederik and Hennig, Philipp and Balles, Lukas},
  journal={Advances in neural information processing systems},
  volume={32},
  year={2019}
}

@article{cardoso2022monai,
  title={Monai: An open-source framework for deep learning in healthcare},
  author={Cardoso, M Jorge and Li, Wenqi and Brown, Richard and Ma, Nic and Kerfoot, Eric and Wang, Yiheng and Murrey, Benjamin and Myronenko, Andriy and Zhao, Can and Yang, Dong and others},
  journal={arXiv preprint arXiv:2211.02701},
  year={2022}
}

@article{kumar2019verified,
  title={Verified uncertainty calibration},
  author={Kumar, Ananya and Liang, Percy S and Ma, Tengyu},
  journal={Advances in neural information processing systems},
  volume={32},
  year={2019}
}

@inproceedings{zhang2021dodnet,
  title={Dodnet: Learning to segment multi-organ and tumors from multiple partially labeled datasets},
  author={Zhang, Jianpeng and Xie, Yutong and Xia, Yong and Shen, Chunhua},
  booktitle={Proceedings of the IEEE/CVF conference on computer vision and pattern recognition},
  pages={1195--1204},
  year={2021}
}

@inproceedings{chen2024test,
  title={Test-Time Medical Image Segmentation Using CLIP-Guided SAM Adaptation},
  author={Chen, Haotian and Xu, Yonghui and Xu, Yanyu and Zhang, Yixin and Cui, Lizhen},
  booktitle={2024 IEEE International Conference on Bioinformatics and Biomedicine (BIBM)},
  pages={1866--1873},
  year={2024},
  organization={IEEE}
}

@article{yeung2023calibrating,
  title={Calibrating the dice loss to handle neural network overconfidence for biomedical image segmentation},
  author={Yeung, Michael and Rundo, Leonardo and Nan, Yang and Sala, Evis and Sch{\"o}nlieb, Carola-Bibiane and Yang, Guang},
  journal={Journal of Digital Imaging},
  volume={36},
  number={2},
  pages={739--752},
  year={2023},
  publisher={Springer}
}

@article{zhang2023threshold,
  title={Threshold-consistent margin loss for open-world deep metric learning},
  author={Zhang, Qin and Xu, Linghan and Tang, Qingming and Fang, Jun and Wu, Ying Nian and Tighe, Joe and Xing, Yifan},
  journal={arXiv preprint arXiv:2307.04047},
  year={2023}
}

@inproceedings{kamnitsas2017unsupervised,
  title={Unsupervised domain adaptation in brain lesion segmentation with adversarial networks},
  author={Kamnitsas, Konstantinos and Baumgartner, Christian and Ledig, Christian and Newcombe, Virginia and Simpson, Joanna and Kane, Andrew and Menon, David and Nori, Aditya and Criminisi, Antonio and Rueckert, Daniel and others},
  booktitle={Information Processing in Medical Imaging: 25th International Conference, IPMI 2017, Boone, NC, USA, June 25-30, 2017, Proceedings 25},
  pages={597--609},
  year={2017},
  organization={Springer}
}

@inproceedings{kim2022fisher,
  title={Fisher sam: Information geometry and sharpness aware minimisation},
  author={Kim, Minyoung and Li, Da and Hu, Shell X and Hospedales, Timothy},
  booktitle={International Conference on Machine Learning},
  pages={11148--11161},
  year={2022},
  organization={PMLR}
}

@article{kull2019beyond,
  title={Beyond temperature scaling: Obtaining well-calibrated multi-class probabilities with dirichlet calibration},
  author={Kull, Meelis and Perello Nieto, Miquel and K{\"a}ngsepp, Markus and Silva Filho, Telmo and Song, Hao and Flach, Peter},
  journal={Advances in neural information processing systems},
  volume={32},
  year={2019}
}

@article{wang2024dual,
  title={Dual-reference source-free active domain adaptation for nasopharyngeal carcinoma tumor segmentation across multiple hospitals},
  author={Wang, Hongqiu and Chen, Jian and Zhang, Shichen and He, Yuan and Xu, Jinfeng and Wu, Mengwan and He, Jinlan and Liao, Wenjun and Luo, Xiangde},
  journal={IEEE Transactions on Medical Imaging},
  year={2024},
  publisher={IEEE}
}

@article{barragan2022towards,
  title={Towards a safe and efficient clinical implementation of machine learning in radiation oncology by exploring model interpretability, explainability and data-model dependency},
  author={Barrag{\'a}n-Montero, Ana and Bibal, Adrien and Dastarac, Margerie Huet and Draguet, Camille and Valdes, Gilmer and Nguyen, Dan and Willems, Siri and Vandewinckele, Liesbeth and Holmstr{\"o}m, Mats and L{\"o}fman, Fredrik and others},
  journal={Physics in Medicine \& Biology},
  volume={67},
  number={11},
  pages={11TR01},
  year={2022},
  publisher={IOP Publishing}
}

@inproceedings{jungo2019assessing,
  title={Assessing reliability and challenges of uncertainty estimations for medical image segmentation},
  author={Jungo, Alain and Reyes, Mauricio},
  booktitle={Medical Image Computing and Computer Assisted Intervention--MICCAI 2019: 22nd International Conference, Shenzhen, China, October 13--17, 2019, Proceedings, Part II 22},
  pages={48--56},
  year={2019},
  organization={Springer}
}

@article{kohl2018probabilistic,
  title={A probabilistic u-net for segmentation of ambiguous images},
  author={Kohl, Simon and Romera-Paredes, Bernardino and Meyer, Clemens and De Fauw, Jeffrey and Ledsam, Joseph R and Maier-Hein, Klaus and Eslami, SM and Jimenez Rezende, Danilo and Ronneberger, Olaf},
  journal={Advances in neural information processing systems},
  volume={31},
  year={2018}
}

@article{yu2024heterogeneity,
  title={Heterogeneity and predictors of the effects of AI assistance on radiologists},
  author={Yu, Feiyang and Moehring, Alex and Banerjee, Oishi and Salz, Tobias and Agarwal, Nikhil and Rajpurkar, Pranav},
  journal={Nature Medicine},
  volume={30},
  number={3},
  pages={837--849},
  year={2024},
  publisher={Nature Publishing Group US New York}
}

@article{isensee2021nnu,
  title={nnU-Net: a self-configuring method for deep learning-based biomedical image segmentation},
  author={Isensee, Fabian and Jaeger, Paul F and Kohl, Simon AA and Petersen, Jens and Maier-Hein, Klaus H},
  journal={Nature methods},
  volume={18},
  number={2},
  pages={203--211},
  year={2021},
  publisher={Nature Publishing Group}
}

@article{liew2022large,
  title={A large, curated, open-source stroke neuroimaging dataset to improve lesion segmentation algorithms},
  author={Liew, Sook-Lei and Lo, Bethany P and Donnelly, Miranda R and Zavaliangos-Petropulu, Artemis and Jeong, Jessica N and Barisano, Giuseppe and Hutton, Alexandre and Simon, Julia P and Juliano, Julia M and Suri, Anisha and others},
  journal={Scientific data},
  volume={9},
  number={1},
  pages={320},
  year={2022},
  publisher={Nature Publishing Group UK London}
}

@article{maier2017isles,
  title={ISLES 2015-A public evaluation benchmark for ischemic stroke lesion segmentation from multispectral MRI},
  author={Maier, Oskar and Menze, Bjoern H and Von der Gablentz, Janina and H{\"a}ni, Levin and Heinrich, Mattias P and Liebrand, Matthias and Winzeck, Stefan and Basit, Abdul and Bentley, Paul and Chen, Liang and others},
  journal={Medical image analysis},
  volume={35},
  pages={250--269},
  year={2017},
  publisher={Elsevier}
}

@article{bennstrom2021automated,
  title={Automated 3D Bone Segmentation using Deep Learning in Scoliosis},
  author={Bennstr{\"o}m, Andreas and Winzell, Filip},
  journal={Master’s Theses in Mathematical Sciences},
  year={2021}
}

\appendix

\section{Appendix}

\subsection{FIP penalty.}
\label{appendix:FIP}
Let $z=f_\theta(x)$ and $p(y \mid z;\phi)$ be the decoder predictive distribution. 
For a small perturbation $\delta \sim \mathcal{N}(0,\sigma^2 I)$, a Taylor expansion yields
\begin{equation}
\mathrm{KL}\!\left(p(\cdot\mid z;\phi)\,\|\,p(\cdot\mid z+\delta;\phi)\right) 
\;\approx\; \tfrac{1}{2}\,\delta^\top I_z \,\delta,
\end{equation}
where
\begin{equation}
I_z := \mathbb{E}_{y \sim p(\cdot \mid z)}\!\left[ \nabla_z \log p(y\mid z;\phi)\,\nabla_z \log p(y\mid z;\phi)^\top \right].
\end{equation}
Taking expectation over $\delta$ gives
\begin{equation}
\mathbb{E}_{\delta}\big[\mathrm{KL}(\cdot)\big] \approx \tfrac{\sigma^2}{2}\,\mathrm{tr}(I_z)
= \tfrac{\sigma^2}{2}\,\mathbb{E}\big[\|\nabla_z \log p(y\mid z;\phi)\|_2^2\big].
\end{equation}
Since $\nabla_z \log p(y\mid z;\phi) = - \nabla_z \mathcal{L}_{\mathrm{task}}(x;\phi)$, the resulting penalty is
\begin{equation}
\mathcal{L}_{\mathrm{FIP}} \propto \mathbb{E}_x\!\left[\|\nabla_z \mathcal{L}_{\mathrm{task}}(x;\phi)\|_2^2\right].
\end{equation}

We use the \emph{empirical Fisher} (evaluated at observed labels) as a computationally efficient approximation of the expected Fisher. While the empirical Fisher is biased, prior work (e.g., Kunstner et al.) shows it is effective for curvature regularization.
\begin{itemize}
  \item If you compute the Fisher w.r.t.\ features \(z\) rather than parameters \(\theta\), add the chain-rule derivation:
  \[
  I_\theta(\theta) \approx J_\theta^z(\theta)^\top\, I_z(\theta)\, J_\theta^z(\theta),
  \]
  where $J_\theta^z$ is the Jacobian of $z$ w.r.t.\ $\theta$; this clarifies how feature-wise Fisher penalization maps to parameter-space stability.
  \item Provide a short empirical remark (or small experiment) showing that the empirical Fisher $\widehat I$ used in training is numerically close (or how it scales) to the population Fisher on held-out validation slices this helps close the theory→practice gap.
\end{itemize}
\section{Theoretical details for Fisher-based regularization}
\label{appendix:theory}

\subsection{Notation and assumptions}
Let $\mathcal{X}$ denote inputs and $\mathcal{Y}$ outputs. Let $p_\theta(y\mid x)$ be a parametric predictive distribution with parameter $\theta\in\mathbb{R}^d$. Denote the \emph{score} by
\[
s_\theta(y;x) \;=\; \nabla_\theta \log p_\theta(y\mid x),
\]
the \emph{per-input Fisher} by
\[
I_x(\theta) \;=\; \mathbb{E}_{Y\sim p_\theta(\cdot\mid x)}\big[\, s_\theta(Y;x)\, s_\theta(Y;x)^\top \,\big],
\]
and the population Fisher by
\[
I(\theta)\;=\;\mathbb{E}_{X}[\, I_X(\theta)\,].
\]

We make the following standard regularity assumptions.

\begin{assumption}[Smoothness]
For every $x,y$ the map $\theta\mapsto \log p_\theta(y\mid x)$ is three times continuously differentiable. There exists $M>0$ such that for all $\theta,x,y$ the third derivative satisfies
\[
\| \nabla_\theta^3 \log p_\theta(y\mid x) \|_{\mathrm{op}} \le M.
\]
\end{assumption}

\begin{assumption}[Compactness / boundedness]
There exists $R>0$ such that we consider perturbations $\Delta\theta$ with $\|\Delta\theta\|_2 \le R$, and all expectations below are finite. (This assumption is only to control remainders.)
\end{assumption}

These assumptions are standard for local Taylor expansions of the log-likelihood (see e.g., textbook derivations of the Fisher information).

\subsection{Local KL expansion (second-order approximation)}
\begin{proposition}[Local KL expansion]
\label{prop:local-kl}
Fix $x\in\mathcal{X}$ and $\theta$. For any $\Delta\theta\in\mathbb{R}^d$ with $\|\Delta\theta\|$ sufficiently small,
\[
\mathrm{KL}\!\big( p_\theta(\cdot\mid x)\,\Vert\, p_{\theta+\Delta\theta}(\cdot\mid x) \big)
\;=\; \tfrac{1}{2}\,\Delta\theta^\top I_x(\theta)\,\Delta\theta \;+\; R_x(\Delta\theta),
\]
where the remainder satisfies $|R_x(\Delta\theta)| \le C_x \|\Delta\theta\|^3$ for a constant $C_x$ depending on the third-derivative bound $M$. If Assumption 1 holds uniformly in $x$, the constant may be chosen uniformly (write $C$).
\end{proposition}

\begin{proof}
By Taylor expansion of $t\mapsto \log p_{\,\theta+t\Delta\theta}(y\mid x)$ about $t=0$:
\begin{multline*}
\log p_{\theta+\Delta\theta}(y\mid x)
= \log p_\theta(y\mid x)
+ \Delta\theta^\top s_\theta(y;x) \\
+ \tfrac{1}{2}\Delta\theta^\top H_\theta(y;x)\Delta\theta
+ r_3(y;x;\Delta\theta),
\end{multline*}
where $H_\theta(y;x)=\nabla^2_\theta \log p_\theta(y\mid x)$ and the remainder obeys $|r_3(y;x;\Delta\theta)| \le C' \|\Delta\theta\|^3$ by Assumption~1. Now
\begin{multline*}
\mathrm{KL}\big(p_\theta(\cdot\mid x) \Vert p_{\theta+\Delta\theta}(\cdot\mid x)\big)
= \mathbb{E}_{Y\sim p_\theta(\cdot\mid x)}\Big[ \log p_\theta(Y\mid x) \\ - \log p_{\theta+\Delta\theta}(Y\mid x) \Big] \\
= -\Delta\theta^\top \underbrace{\mathbb{E}[s_\theta(Y;x)]}_{=0}
\;-\; \tfrac{1}{2}\Delta\theta^\top \underbrace{\mathbb{E}[H_\theta(Y;x)]}_{=-I_x(\theta)}\Delta\theta \\
\;-\; \mathbb{E}[r_3(Y;x;\Delta\theta)].
\end{multline*}
Using $\mathbb{E}[s_\theta]=0$ and $\mathbb{E}[H_\theta]=-I_x(\theta)$ (standard under regularity), the identity follows with $R_x(\Delta\theta)=-\mathbb{E}[r_3(Y;x;\Delta\theta)]$ and $|R_x(\Delta\theta)|\le C_x\|\Delta\theta\|^3$.
\end{proof}

\subsection{From KL to predictive-distance and to bounded functionals}
The following chain of standard inequalities turns Fisher-control into a bound on changes in model outputs and thus on bounded Lipschitz functionals of the predictive distribution (accuracy, expected loss under bounded loss, and under additional mild assumptions calibration metrics).

\begin{proposition}[Stability of predictive distribution]
\label{prop:kl-to-l1}
For any $x$ and any $\Delta\theta$,
\[
\| p_\theta(\cdot\mid x) - p_{\theta+\Delta\theta}(\cdot\mid x) \|_{1}
\;\le\; \sqrt{2\;\mathrm{KL}\big( p_\theta(\cdot\mid x) \Vert p_{\theta+\Delta\theta}(\cdot\mid x)\big)}.
\]
Consequently, averaging over $X\sim\mathcal{D}_X$,

\begin{multline*}
\mathbb{E}_X\| p_\theta(\cdot\mid X) - p_{\theta+\Delta\theta}(\cdot\mid X) \|_{1} \\
\;\le\; \sqrt{2\,\mathbb{E}_X\Big[\mathrm{KL}\big( p_\theta(\cdot\mid X) \Vert p_{\theta+\Delta\theta}(\cdot\mid X)\big)\Big]}.
\end{multline*}
\end{proposition}

\begin{proof}
The first line is Pinsker's inequality: $\|p-q\|_1 \le \sqrt{2\,\mathrm{KL}(p\Vert q)}$. The second line follows by applying Jensen / Cauchy--Schwarz to bound the expectation of the pointwise square roots:
\[
\mathbb{E}_X \sqrt{a(X)} \le \sqrt{\mathbb{E}_X a(X)}\quad\text{for }a(X)\ge0.
\]
Set $a(X)=2\,\mathrm{KL}(\cdot)$.
\end{proof}

Combining Proposition~\ref{prop:local-kl} and Proposition~\ref{prop:kl-to-l1} yields the following (local) bound.

\begin{corollary}[Local Fisher $\Rightarrow$ small change in predictions]
\label{cor:fisher-to-l1}
Under Assumptions 1--2, for $\|\Delta\theta\|$ small,
\[
\mathbb{E}_X\| p_\theta(\cdot\mid X) - p_{\theta+\Delta\theta}(\cdot\mid X) \|_{1}
\;\le\; \sqrt{\,\Delta\theta^\top I(\theta)\Delta\theta \;+\; O(\|\Delta\theta\|^3)\, }.
\]
\end{corollary}

\begin{proof}
Apply Proposition~\ref{prop:local-kl} and then Proposition~\ref{prop:kl-to-l1}. Averaging the per-$x$ second-order term produces $\tfrac12\Delta\theta^\top I(\theta)\Delta\theta$ inside the square root; constants combine to give the displayed form.
\end{proof}

\subsection{Implication for bounded Lipschitz functionals (including calibration proxies)}
Let $\Phi$ be a functional that maps the conditional distribution $p(\cdot\mid x)$ to a bounded scalar of interest (e.g., expected 0–1 loss under a fixed decision rule, Brier score, or many empirical calibration proxies). Suppose $\Phi$ is $L$–Lipschitz with respect to the $\ell_1$ norm on the conditional distribution, uniformly in $x$:
\[
\big|\Phi(p(\cdot\mid x))-\Phi(q(\cdot\mid x))\big| \le L\,\|p(\cdot\mid x)-q(\cdot\mid x)\|_1.
\]
Then averaging over $X$ we obtain:

\begin{proposition}[Change in bounded Lipschitz functionals]
\label{prop:lipschitz-functional}
Under the Lipschitz condition above,
\[
\big| \mathbb{E}_X[\Phi(p_\theta(\cdot\mid X))] - \mathbb{E}_X[\Phi(p_{\theta+\Delta\theta}(\cdot\mid X))] \big|
\;\le\; L\ \mathbb{E}_X\| p_\theta(\cdot\mid X) - p_{\theta+\Delta\theta}(\cdot\mid X)\|_1.
\]
Combined with Corollary~\ref{cor:fisher-to-l1}, for small $\|\Delta\theta\|$,
\[
\big|\mathbb{E}_X\Phi(p_\theta)-\mathbb{E}_X\Phi(p_{\theta+\Delta\theta})\big|
\;\le\; L\ \sqrt{\,\Delta\theta^\top I(\theta)\Delta\theta \;+\; O(\|\Delta\theta\|^3)\, }.
\]
\end{proposition}

\begin{remark}
Many commonly used quantities satisfy the Lipschitz condition or can be approximated by Lipschitz functionals on the predictive distribution. For calibration metrics such as ECE computed via a finite number of bins, the map from predictive probabilities to the ECE estimate is Lipschitz (the Lipschitz constant depends on the number of bins and the binning rule). Hence, changes in ECE from small parameter perturbations are similarly controlled by the Fisher-based quantity above (up to constants that depend on the binning and estimator).
\end{remark}

\subsection{Discussion and practical interpretation}
\begin{itemize}
  \item {\bf Local regime.} All bounds above are \emph{local} in $\Delta\theta$: they are meaningful when the parameter perturbation is small enough for the Taylor expansion remainder to be negligible. Thus statements that ``smaller Fisher implies better robustness under arbitrary large changes'' are too strong; we claim a local stability guarantee: penalizing Fisher reduces the \emph{local} sensitivity of the predictive distribution to parameter perturbations, which in turn bounds changes in bounded Lipschitz quantities (accuracy, bounded losses, calibration proxies) by $\mathcal{O}\big(\sqrt{\Delta\theta^\top I(\theta)\Delta\theta}\big)$.
  \item {\bf Empirical Fisher vs.\ true Fisher.} In practice we compute an empirical Fisher matrix $\widehat I$ from finite training data or even use gradient-squared approximations (empirical Fisher). Under standard concentration results for sample covariance matrices, $\widehat I$ converges to $I$ as sample size grows; finite-sample deviations produce additional error terms in the bounds above. We recommend the manuscript explicitly note this approximation error and, if possible, report sensitivity of results to sample size and the choice of empirical Fisher estimator.
  \item {\bf Frozen encoder / feature-Fisher.} If the encoder is frozen and the Fisher penalty is computed with respect to intermediate representations $z=f_\theta(x)$ (i.e., \emph{feature-Fisher}), then the same chain of reasoning applies by viewing the decoder parameterization as $p(y\mid z;\phi)$ and controlling the sensitivity of $p(y\mid z)$ to changes in $z$. Under the chain rule, a Fisher penalty on representations has the effect of limiting the magnitude of Jacobian-mediated parameter perturbations (see the remark below and add an explicit derivation if you use feature-Fisher in the implementation).
\end{itemize}

\end{document}